%% file: main.tex
\documentclass{article} 
\usepackage{collas2026_conference,times}
\usepackage{graphicx}


\input{math_commands.tex}

\usepackage{hyperref}
\hypersetup{
    colorlinks=true,
    linkcolor=red,
    filecolor=magenta,
    urlcolor=blue,
    citecolor=purple,
    pdftitle={Towards Continual Motion-Language Agents},
    pdfpagemode=FullScreen,
}

\newcommand{\MethodOLoRAMoE}{\textsc{O-LoRA-MoE}}
\newcommand{\MethodLoRAMoE}{\textsc{LoRA-MoE}}

\title{Towards Continual Motion-Language Agents: \\ 
LoRA Variants for Incremental Motion Understanding and Generation}

\author{Bertram Taetz $^*$\\
  IT \& Engineering \\
  International University of Applied Sciences \\
  Erfurt, Germany\\
  \texttt{bertram.taetz@iu.org} \\
   \And Hugo Albuquerque Cosme da Silva \thanks{These authors contributed equally to this work.} \\
  IT \& Engineering \\
  International University of Applied Sciences \\
  Erfurt, Germany\\
  \texttt{hugo.silva@iu-study.org} \\
  \And Gabriele Bleser-Taetz\\
  IT \& Engineering \\
  International University of Applied Sciences \\
  Erfurt, Germany\\
  \texttt{gabriele.bleser-taetz@iu.org} 
}

\collasfinalcopy

\begin{document}
\maketitle
\lhead{Updated author version of the CoLLAs 2026 paper}

\begingroup
\small
\noindent\textbf{Version note.} This manuscript is an updated author version of the paper published at the Fifth Conference on Lifelong Learning Agents (CoLLAs 2026). Relative to the published camera-ready version, it adds analysis and comparative evaluations of merged-adapter routing, clarifies the methodological description and interpretation of results, and integrates the approach overview into the main text.
\par
\endgroup

\begin{abstract}
Motion-language agents must possess the bidirectional capability to both understand human movement (motion-to-text, M2T) and generate it from natural language (text-to-motion, T2M). While foundational models have achieved strong performance in static settings, autonomous agents operating in dynamic environments must continuously incorporate new motion concepts - such as novel athletic styles or specialized gestures - without catastrophic forgetting of previously acquired skills. We investigate the stability-plasticity trade-off in bidirectional motion-language learning under sequential task exposure.
Building on a frozen large language model backbone, we introduce low-rank adaptation (LoRA) variants designed to mitigate inter-task interference. We specifically propose  mixture-of-experts architectures that utilize an autoencoder-based router to select task-specific  experts at inference time, so that no task-label is needed. To evaluate these methods, we establish an initial reproducible five-task benchmark derived from HumanML3D through semantic clustering of motion descriptions.
Our experimental results demonstrate  near-zero forgetting across both M2T and T2M directions while maintaining high generation and captioning quality. Furthermore, we show that hard expert selection via routing can significantly outperform soft expert blending, via logit-mixture, in quality metrics, indicating that preserving mixture consistency is critical for maintaining performance in our continual learning setting. Finally, we observe that a divergence between token-level accuracy and downstream generation quality may occur, highlighting the need for more comprehensive evaluation protocols in future research on lifelong motion-language agents. Our code and the benchmark construction are available under \url{https://github.com/bertramtaetz/towards-cl-language-motion-agents}.

\end{abstract}

\input{sections/introduction}

\input{sections/related_work}
\input{sections/method}

\input{sections/results}

\input{sections/discussion}

\input{sections/conclusion}


\bibliography{collas2026_conference}
\bibliographystyle{collas2026_conference}

\input{sections/appendix}

\end{document}

%% file: math_commands.tex
\usepackage{amsmath,amsfonts,bm}

\def\eqref#1{equation~\ref{#1}}

\def\1{\bm{1}}

\DeclareMathAlphabet{\mathsfit}{\encodingdefault}{\sfdefault}{m}{sl}
\SetMathAlphabet{\mathsfit}{bold}{\encodingdefault}{\sfdefault}{bx}{n}



%% file: sections/introduction.tex
\section{Introduction}

Human motion is a rich, high-dimensional signal central to human-computer, machine-, or robot-interaction, serving as the foundation for activity understanding and response generation in both virtual and physical agents. Recent advancements have demonstrated that learning a shared representation between motion and natural language enables powerful bidirectional capabilities, specifically text-to-motion (T2M) generation and motion-to-text (M2T) captioning. As demonstrated by the Motion-Agent framework~\citep{wu_motion-agent_2024}, these capabilities can be suitably combined with reasoning models like GPT-4 to generate and customize highly complex motion sequences through multi-turn conversations. This combination was enabled by the MotionLLM that bridges the gap between motion and text~\citep{wu_motion-agent_2024}, via T2M and M2T. However, most existing systems for motion understanding and motion generation are trained in static settings with all data available upfront, whereas a truly capable agent operating "in the wild" must continuously incorporate new concepts - such as novel athletic styles, domain-specific gestures, or specialized robotic manipulation skills - without retraining from scratch.
The demand for such motion-language agents is driven by the need for autonomous entities - including humanoid robots, digital twins, or virtual assistants - to interact naturally in dynamic environments. This requirement introduces the challenge of continual learning (CL), where an agent must learn from a non-stationary task stream while avoiding catastrophic forgetting. This challenge is particularly amplified in bidirectional motion-language agents because the M2T and T2M directions often share parameters and can interfere, and motion domains can differ substantially in both kinematics and semantics.
Our work is motivated by the real-world deployment constraints of Task-Incremental Learning with Task-Agnostic Inference. It is motivated by scenarios like the following:
\begin{itemize}
\item{
    Embodied Skill Acquisition: A domestic robot may sequentially learn to "Walk" and then coordinate "Arms/Hands," but it must decide which skill to activate during interaction based on unlabeled verbal or visual cues.
    }
\item{
    Personalized Interactive Agents: Virtual characters in AR/VR must learn new interaction behaviors as sequential tasks, while activating those during open interaction.}
\item{
    Computational Efficiency: Retraining large-scale foundation models, every time a new motion concept is introduced is computationally prohibitive.}
\end{itemize}

To address these issues, we focus on parameter-efficient continual learning. Instead of updating all parameters, we utilize compact low-rank adapters (LoRA) and also investigate Orthogonal LoRA (O-LoRA) variants that impose a geometric constraint to reduce inter-task interference by isolating updates into near-orthogonal subspaces. Specifically, we propose LoRA-MoE and O-LoRA-MoE, which are  task-label-free (at inference) mixture-of-experts architectures that utilize an autoencoder-based router to select task-specific experts at inference time. These approaches ensure that the model remains scalable while maintaining the isolation necessary to prevent catastrophic interference between distinct kinematic behaviors.
The proposed framework is uniquely aligned with the compositional nature of motion-language modeling, as it utilizes parameter-efficient task-specific experts to allows embodied agents to specialize in new behaviors without the prohibitive cost of full retraining. To evaluate our approach, we establish a reproducible five-task benchmark derived from HumanML3D~\citep{guo2022humanml3d} through semantic clustering of motion descriptions. 
The primary contributions of this paper are:
\begin{itemize}
\item{
    Unified bidirectional continual motion-language setup: We study CL in a single framework including both M2T and T2M under sequential task exposure with a unified evaluation protocol.}
\item{
    LoRA variants for task-incremental learning of T2M and M2T: We assign LoRA adapters to each task and study the effect of different adapter combination strategies, like progressive merging, linear blending of selected adapters (soft routing) and hard selection of adapters (hard routing).
    }
\item{
    Task-label-free  expert routing: We propose an autoencoder-based router to select  LoRA experts, enabling task-label-free inference. Figure~\ref{fig:overview} summarizes the training and inference workflow.
    }
\item{
    Reproducible semantic-clustering benchmark: We introduce a balanced five-task benchmark derived from caption semantics to provide evidence-backed insights for future research in lifelong motion-language agents.}
\end{itemize}

%% file: sections/related_work.tex




\section{Related Work}
\label{sec:related_work}
\paragraph{Motion-language modeling.}
Modeling the relationship between human motion and natural language has seen rapid progress in recent years, driven by advances in large-scale datasets and generative modeling, as extensively documented in recent literature surveys like \citep{zhu_2024_motgensurv}. Early work such as HumanML3D~\citep{guo2022humanml3d} established a benchmark for paired motion-text data, enabling both motion captioning (M2T) and motion generation (T2M). On the modeling side, approaches like MotionCLIP~\citep{tevet2022motionclip} leverage contrastive learning to align motion representations with language embeddings, while TEMOS~\citep{petrovich2022temos} provides a unified framework for bidirectional motion-language mapping. 
More recent methods have focused on improving generation quality and controllability. Autoregressive models such as T2M-GPT~\citep{zhang2023t2mgpt} treat motion as a sequence of discrete tokens, enabling scalable transformer-based generation. Diffusion-based approaches, including MotionDiffuse~\citep{zhang2022motiondiffuse} and MDM~\citep{tevet2023human}, achieve strong performance in terms of realism and diversity by modeling motion distributions in continuous space. As highlighted by \cite{khani_motion_2025}, the field is now trending toward "motion-as-language" foundation models.  MotionGPT~\citep{jiang_motiongpt_2023}, MotionGPT2~\citep{wang_motiongpt-2_2024}, MotionGPT3~\citep{zhu_motiongpt3_2025} and Motion-Agent~\citep{wu_motion-agent_2024} extend these ideas toward unified agentic architectures capable of both understanding and generating motion. Despite these advances, existing methods are typically trained in static offline settings and do not address incremental adaptation.

\paragraph{CL and catastrophic forgetting.} 
Continual learning seeks to empower artificial intelligence (AI) systems to incrementally acquire, update, and accumulate knowledge throughout their lifetime, mimicking the adaptability of biological systems \citep{parisi_continual_2019}. One of the primary obstacles in continual learning with neural networks is catastrophic forgetting \citep{bower_catastrophic_1989}, a phenomenon where adapting to a new data distribution typically results in a dramatic loss of performance on previously learned distributions. This challenge is fundamentally a facet of the stability-plasticity trade-off: excessive learning plasticity interferes with memory stability, and vice-versa \citep{wang_comprehensive_2024,parisi_continual_2019}.
Representative CL methods and their categorization are described in the survey of \cite{wang_comprehensive_2024}.
Recent advancements have further extended CL into more complex scenarios, such as Task-Free Continual Learning (TFCL)~\citep{Aljundi_2019_CVPR}, where task boundaries are unknown, and Online Continual Learning (OCL)~\citep{aljundi2019gradient}, where data arrives in a one-pass stream. Moreover, there is a growing emphasis on representation-based approaches that leverage the robustness of self-supervised learning~\citep{gallardo2021selfsupervised} or large-scale pre-training to mitigate forgetting~\citep{mehta2021empirical}. Optimization-based methods have also gained traction, explicitly manipulating gradient projections \citep{saha2020gpm}. 
These perspectives are particularly relevant to the study of motion–language agents, where high-dimensional temporal dynamics and bidirectional mappings necessitate solutions that balance both intra-task and inter-task generalizability.

\paragraph{Parameter-efficient adaptation and CL.} 
Parameter-efficient fine-tuning (PEFT) methods have emerged as a scalable alternative to full model retraining. Low-Rank Adaptation (LoRA)~\citep{hu2021lora} introduces trainable low-rank updates to pretrained weights, significantly reducing the number of trainable parameters. Related approaches include adapter layers~\citep{houlsby2019parameter} and prefix tuning~\citep{li2021prefix}, which enable modular and efficient task adaptation. These methods are particularly attractive in continual learning settings, where repeated full fine-tuning is computationally prohibitive.  The synergy between these methods and CL has given rise to the paradigm of Parameter-Efficient Continual Fine-Tuning (PECFT)~\citep{coleman_parameter-efficient_2025}. This approach addresses the stability-plasticity dilemma by freezing the foundational pre-trained backbone to preserve general-purpose representations while introducing compact, task-specific modules for incremental adaptation.
Building on this, recent research has focused on interference prevention through explicit architectural design. For instance, Orthogonal LoRA (O-LoRA)~\citep{ortholora2023} and InfLoRA~\citep{liang_inflora_2024} enforce mathematical orthogonality between task-specific updates, ensuring that new learning occurs in the null space of previous tasks to achieve exceptional stability. 
Beyond individual adapters, the field is moving toward compositional isolation and merging strategies to handle task-free inference and scalability. Mixture-of-LoRA-Experts (MoLE)~\citep{wu2024mixture} introduces hierarchical frameworks that learn layer-wise gating weights to compose multiple LoRA experts without base model retraining. To standardize diverse strategies, unified frameworks like LAE (Learning-Accumulation-Ensemble)~\citep{Gao_2023_ICCV} and APER (AdaPt and mERge)~\citep{zhou2024revisiting} have been proposed to decouple the specific PEFT module from the CL strategy, allowing for flexible experimentation across different model architectures. These advancements are particularly critical for motion-language agents, where preserving expert isolation in a multimodal, bidirectional context is essential for maintaining both generation quality and captioning accuracy.

\paragraph{Multimodal CL and LLM-based agents}
Extending CL to multimodal settings introduces significant challenges, primarily due to cross-modal interference and the risk of alignment drift~\citep{guo_continual_2025}. A recent focus in this area is Continual Multimodal Instruction Tuning (CMIT), which aims to efficiently adapt Multimodal Large Language Models (MLLMs) to sequential tasks while preserving both general and task-specific knowledge~\citep{guo_continual_2025}. Traditional approaches often utilize fixed architectures that suffer from task architecture conflict - where different layers exhibit varying sensitivities to different tasks - and modality imbalance, where dominant modalities suppress weaker ones during training. To address these issues, the D-MoLE (Dynamic Mixture of Curriculum LoRA Experts) framework proposes an architecture-evolution approach~\citep{dmole2025}. It employs a dynamic layer-wise expert allocator guided by training-free zero-cost proxies to identify critical layers for adaptation, alongside a gradient-based inter-modal continual curriculum that dynamically adjusts update ratios between language and modality-specific encoders to ensure balanced learning.
Beyond the adaptation of static models, the field is rapidly evolving toward lifelong learning LLM-based agents~\citep{zheng_lifelong_2026}. Unlike standard MLLMs that process fixed data distributions, these agents are autonomous entities capable of interacting with ever-changing environments. Their architecture is typically categorized into three synergistic modules: perception for multimodal input integration, memory for storing evolving knowledge, and action for grounded environmental interaction. The memory module comprising working, episodic, semantic, and parametric memory, was proposed for developing coherent long-term behavior and mitigating catastrophic forgetting~\citep{zheng_lifelong_2026}. Recent progress in Vision-Language-Action (VLA) models, such as iManip~\citep{zheng2025imanip} and LEGION~\citep{meng2025preserving}, further underscores the necessity of incremental adaptation in embodied agents, where models must continuously acquire new manipulation skills while preserving prior multimodal knowledge. Our work contributes to multimodal continual learning via the language- and motion domain and explicitly considers both M2T and T2M tasks under sequential learning.

%% file: sections/method.tex
\section{Methods}
\label{sec:method}

\begin{figure}[t]
    \centering
    \includegraphics[width=0.95\textwidth]{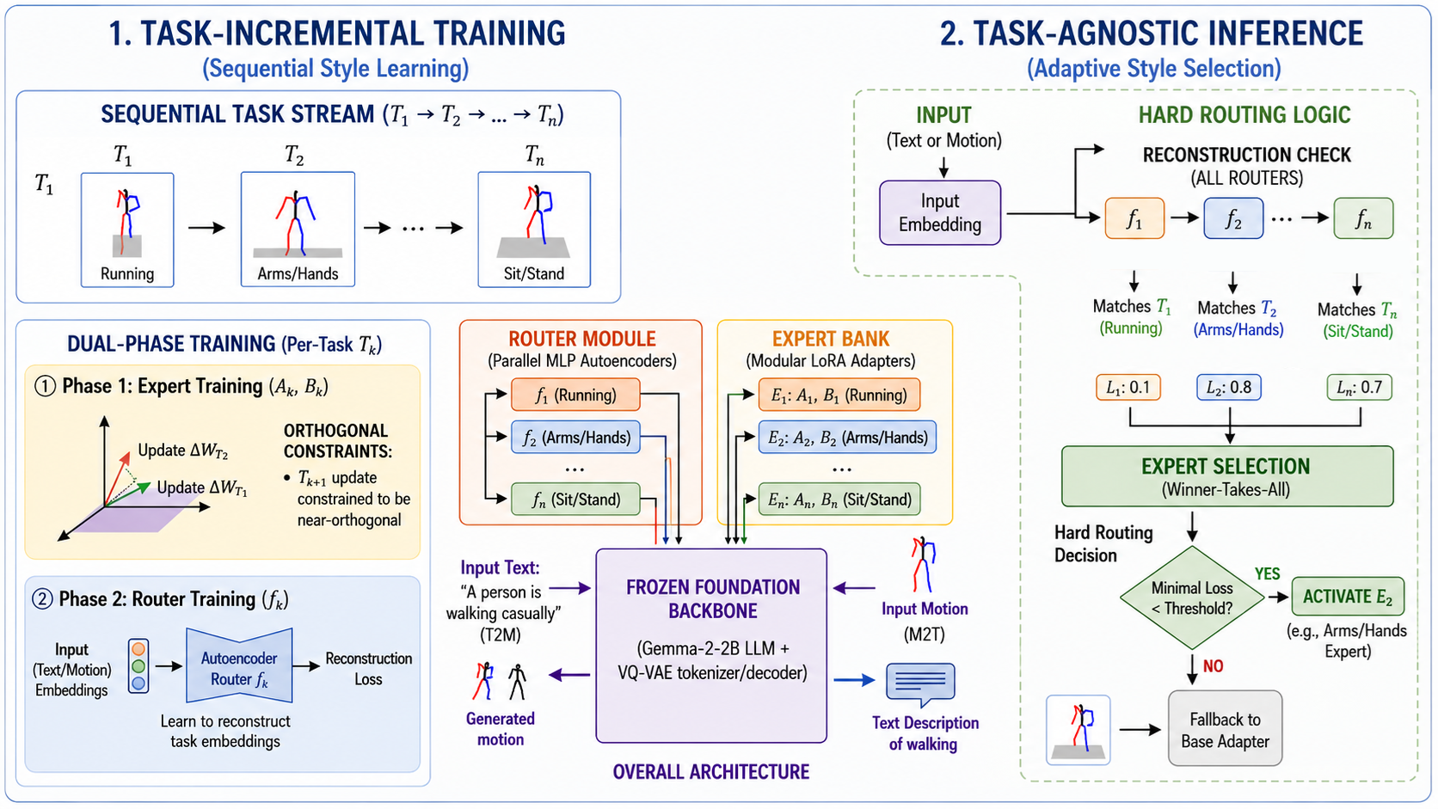}
    \caption{Overview of \MethodOLoRAMoE{} for continual bidirectional motion-language learning. \textbf{Left:} sequential training of task-specific experts and autoencoder routers. \textbf{Center:} the frozen backbone, router module, and expert bank supporting T2M and M2T. \textbf{Right:} task-label-free hard expert selection with direction-specific fallback.}
    \label{fig:overview}
\end{figure}

\paragraph{Approach overview.}
Figure~\ref{fig:overview} illustrates our continual motion-language approach using \MethodOLoRAMoE{}.
For each new task, we first train a task-specific LoRA expert and then an autoencoder router on mean-pooled LLM input embeddings, while keeping the backbone and previously learned experts fixed.
At inference, the routers evaluate the input without task labels, and the expert with the lowest reconstruction error is selected.
If all routers exceed their respective thresholds, the model falls back to the direction-specific base adapter.
This workflow applies to both text-to-motion and motion-to-text.
\MethodLoRAMoE{} omits the orthogonality regularizer, while the Top-$K$ variants combine multiple selected experts.

In the following, we first discuss the backbone model used in this work and its directional pretraining (Section \ref{sec:transfer_motionllm}). Then, we formalize the specific CL problem (Section \ref{sec:problem_setup}).
The considered CL approaches are presented in Sections \ref{sec:method_transfer} to \ref{sec:olora_moe_m}. {\bf Transfer} represents the baseline via sequential fine-tuning, {\bf O-LoRA-PM} uses a progressively merged orthogonal LoRA approach, {\bf \MethodOLoRAMoE{}} uses a hard routed mixture of experts, and {\bf \MethodOLoRAMoE{}-$K$} use a Top$K$ mixture of experts (with soft weights over the routed Top$K$ set ($K>1$)). We also compare to {\bf \MethodLoRAMoE{}} and {\bf \MethodLoRAMoE{}-$K$} without orthogonalization.
Finally, Section \ref{sec:multitask} presents the upper bound reference via joined multi-task training ({\bf Multi-task}).

\subsection{Motion-Agent base model (MotionLLM)}
\label{sec:transfer_motionllm}

We build on the static Motion-Agent backbone from \cite{wu_motion-agent_2024}, called \texttt{MotionLLM}. It uses the Gemma-2-2B backbone \citep{mesnard2024gemma} and combines (i) a frozen VQ-VAE motion tokenizer and decoder, and (ii) a decoder-only causal LLM whose vocabulary is extended to represent motion codes.
Given an input motion sequence, the VQ-VAE tokenizer maps it to a discrete code sequence.
The backbone tokenizer is then extended with special tokens \texttt{<Motion>}, \texttt{</Motion>}, and a set of motion code tokens \texttt{<Motion\_i>}.
These additional tokens are appended to the LLM vocabulary and allow the model to generate and consume motion as ordinary discrete symbols.

MotionLLM supports T2M (input: caption, target: motion-token sequence delimited by \texttt{<Motion>} and \texttt{</Motion>}) and M2T (input: motion-token sequence inside \texttt{<Motion>} and \texttt{</Motion>}, target: caption) using direction-specific instruction templates and LoRA adapters.
In both cases, the model is trained with teacher-forced causal language modeling, but only the response region contributes to the loss (prompt tokens are masked).

\paragraph{Pretrained direction adapters and frozen interface.}
The Gemma backbone is extended with one pretrained LoRA adapter per direction (T2M and M2T).
These direction adapters function as basic ``motion token interfaces''. They are trained on a small portion of data (holdout data) as described in Appendix \ref{sec:benchmark_construction}.
We train the direction-specific LoRA adapters using the same teacher-forced causal language modeling objective as used in \eqref{eq:transfer_task_loss}.
The goal is to learn a minimal prior for interacting with the extended motion-token vocabulary (including motion token embeddings).
At the start of CL, the pretrained adapter for the considered direction is merged into the backbone weights.
This preprocessing step is needed because otherwise early CL stages would spend a substantial fraction of their adaptation budget on learning the motion-token interface itself, conflating interface learning with task adaptation.



\subsection{CL problem setup and notation}
\label{sec:problem_setup}
We consider a bidirectional motion-language agent that encounters a sequential stream of $T$ tasks, $\mathcal{S} = \{\mathcal{T}_1, \mathcal{T}_2, \dots, \mathcal{T}_T\}$. Each task $\mathcal{T}_k$ is defined by a dataset $\mathcal{D}_k = \{(x_i, m_i)\}_{i=1}^{n_k}$ containing pairs of natural language descriptions $x$ and discretized motion token sequences $m$. The agent is trained to perform two primary functions: M2T, mapping $m \to x$, and T2M, mapping $x \to m$.

To accurately measure the agent's evolution, we distinguish between the training stage $s \in \{1, \dots, T\}$ and the task label $k \in \{1, \dots, T\}$. The stage $s$ represents the temporal progress of the model. Specifically, at stage $s$, the agent has completed training on tasks $\mathcal{T}_1$ through $\mathcal{T}_s$. The task label $k$ identifies the specific semantic distribution (e.g., ``Running'' vs. ``Jumping'') being addressed. In our benchmark, training at stage $s$ is conducted exclusively on dataset $\mathcal{D}_k$ where $k=s$.
This corresponds to a \textbf{Task-Incremental Learning (TIL)} protocol. At each stage $s$, the model has access to the data $\mathcal{D}_k$ (for $k=s$). 

We evaluate the agent under \textbf{Task-Agnostic Inference}. Unlike standard TIL, which assumes an ``oracle'' provides the task identity $k$ at test time to select the correct parameters, our model must process an input (text prompt or motion sequence) without a task label. 
\subsection{Transfer (Sequential Fine-Tuning)}
\label{sec:method_transfer}
Transfer (learning) is the standard sequential fine-tuning baseline for continual learning. The model is trained on tasks one after another, reusing the same parameterization for all tasks.
In our setting, adaptation is performed with a single LoRA adapter per direction (T2M or M2T) that is updated continuously across the task stream.
Because the same adapter weights are overwritten by later tasks, this baseline typically exhibits strong catastrophic forgetting and serves as a lower bound.

We use the Motion-Agent backbone with the respective merged pretrained direction adapters, as described in Section \ref{sec:transfer_motionllm}.
LoRA injects a low-rank update into selected linear projections:
\begin{equation}
\label{eq:transfer_lora}
W = W_0 + \Delta W, \qquad \Delta W = \tfrac{\alpha}{r}BA,
\end{equation}
where $W_0$ is frozen, $A\in\mathbb{R}^{r\times d_{\text{in}}}$ and $B\in\mathbb{R}^{d_{\text{out}}\times r}$ are trainable, $r$ is the rank, and $\alpha$ is a scaling factor.
The target module set targets only the query and value projections for all approaches.

At training stage $s$, the model continues training the same direction adapter (shared across all tasks) on dataset $\mathcal{D}_k$ (with $k$=$s$) for a fixed number of epochs. No explicit forgetting mitigation is applied in this approach.

For both directions (T2M and M2T), the model is trained with a teacher-forced causal language modeling loss.
Given an input-target token sequence $\mathbf{z}=(z_1,\ldots,z_S)$ and a supervised index set $\Omega$, the task loss is
\begin{equation}
\label{eq:transfer_task_loss}
\mathcal{L}_{\text{task}}^{(t)}
= - \frac{1}{|\Omega|}\sum_{i\in\Omega} \log p_{\theta,\phi}(z_i\mid z_{<i}),
\end{equation}
where $\theta$ denotes the (merged) backbone parameters and $\phi$ denotes the shared LoRA adapter parameters for the chosen direction.
A more elaborate explanation of the task loss for the specific directions can be found in Appendix \ref{sec:taskloss_explained}.  

At test time, per direction, a single adapter is used for all tasks.
This makes deployment task-label-free, but also means that later tasks can overwrite earlier-task competence.

\subsection{O-LoRA-PM (Progressive Merge)}
\label{sec:progressive_merge}
O-LoRA-PM follows the O-LoRA training objective, i.e. masked causal language modeling with an orthogonality penalty~\citep{ortholora2023} and represents an orthogonal low-rank adaptation baseline. It merges the newly learned adapter into the backbone after each task and discards the adapter modules. This yields a single continuously updated
model (task-label-free at test time) while retaining O-LoRA’s orthogonality constraint during training.

At training stage $s$, O-LoRA-PM instantiates a fresh LoRA adapter 
and optimizes only that adapter while the backbone is frozen. 
To mitigate interference, the method uses an orthogonality regularizer as proposed in \citep{ortholora2023}.
Let $A_{k,\ell}\in\mathbb{R}^{r\times d_{\text{in}}}$ denote the LoRA down-projection matrix of the current task adapter (task $k$) at transformer module/layer $\ell$.
The implementation stores frozen copies of past $A$ matrices and computes cross-task overlap terms
\begin{equation}
\label{eq:olora_full_overlap}
O_{j,k,\ell} = A_{j,\ell}A_{k,\ell}^{\top} \in \mathbb{R}^{r\times r},
\qquad
\mathcal{L}_{\text{orth}}^{(k)} = \frac{1}{N}\sum_{\ell}\sum_{j<k} \lVert O_{j,k,\ell}\rVert_F^2,
\end{equation}
where $\lVert\cdot\rVert_F$ is the Frobenius norm and $N$ is the number of accumulated $(j,\ell)$ terms.
For task $k$, the adapter is trained by minimizing the following total loss
\begin{equation}
\label{eq:olora_full_total_loss}
\mathcal{L}_{\text{total}}^{(k)} = \mathcal{L}_{\text{task}}^{(k)} + \lambda_{\text{orth}}\,\mathcal{L}_{\text{orth}}^{(k)},
\end{equation}
with default $\lambda_{\text{orth}}=0.5$. Note, $\mathcal{L}_{\text{task}}$ is given in \eqref{eq:transfer_task_loss}.

\subsection{\MethodLoRAMoE{} and \MethodOLoRAMoE{} (Mixture-of-Experts, hard routing)}
\label{sec:olora_moe}
\MethodLoRAMoE~and \MethodOLoRAMoE{} are our proposed mixture-of-experts variants for low-rank adaptation.
\MethodOLoRAMoE{} combines (i) per-task LoRA experts trained with the O-LoRA orthogonality regularizer and (ii) an autoencoder-based router that selects an expert at inference time based on reconstruction error, enabling task-agnostic deployment. \MethodLoRAMoE~works analogously, but without orthogonality regularizer during training.  

\subsubsection{Task-incremental training}
\label{sec:moe_training}
At training stage $s$, we perform two consecutive optimization phases for current task $k$ (with $k=s$):
(1) train the new LoRA expert $\phi_k$ on task $\mathcal{T}_k$, using the orthogonality regularizer against prior experts for \MethodOLoRAMoE{} and setting $\lambda_{\mathrm{orth}}=0$ for \MethodLoRAMoE{}, and
(2) train the router component corresponding to expert $k$ on embeddings extracted from $\mathcal{T}_k$. Note, in contrast to \textsc{O-LoRA-PM}, the LoRA experts ($\phi_k$) are not progressively merged into the backbone model, but selected via the router as described below. The losses are the same as in Section \ref{sec:progressive_merge}.

\paragraph{Router embedding extraction.}
After expert training, we train a router for task-agnostic inference.
We represent each sample by an embedding $\mathbf{e}\in\mathbb{R}^{H}$ in the LLM hidden space, constructed by mean-pooling LLM input embeddings.
The embedding source depends on direction:
\begin{itemize}
  \item \textbf{M2T:} $\mathbf{e}$ is the mean of the motion-token input embeddings, computed from the (VQ-VAE) motion token sequence mapped into the extended LLM vocabulary.
  \item \textbf{T2M:} the router operates on mean-pooled LLM input embeddings. These embeddings are derived from the text prompt before any motion tokens are generated. The input dimension for these embeddings ($\mathbf{e}$) is $2304$, which corresponds to the hidden size of the Gemma-2-2B backbone~\citep{mesnard2024gemma} used for the encoding.
\end{itemize}

\paragraph{Router training loss.}
The router consists of one lightweight MLP autoencoder $f_k$ per task/expert $k$.
When learning task $k$, we freeze all prior autoencoders $\{f_j\}_{j<k}$ and train only $f_k$ by minimizing the mean-squared reconstruction error over collected embeddings $\mathcal{E}_k$:
\begin{equation}
\label{eq:moe_router_loss}
\mathcal{L}_{\text{router}}^{(k)} = \frac{1}{|\mathcal{E}_k|}\sum_{\mathbf{e}\in\mathcal{E}_k} \lVert f_k(\mathbf{e}) - \mathbf{e} \rVert_2^2.
\end{equation}

\paragraph{Thresholding for unseen detection.}
\label{sec:unseendet}
After training $f_k$, we compute per-sample reconstruction losses $\ell_k(\mathbf{e})=\lVert f_k(\mathbf{e})-\mathbf{e}\rVert_2^2$ on the embeddings used for router training.
We store a per-expert threshold
\begin{equation}
\label{eq:moe_threshold}
\tau_k = \mu_k + \kappa\,\sigma_k, \qquad \kappa=2.0,
\end{equation}
where $\mu_k$ and $\sigma_k$ denote the mean and standard deviation of $\ell_k(\mathbf{e})$.
At inference time, this enables an unseen decision when all experts exceed their thresholds.

\subsubsection{Task-agnostic inference: hard routing and fallback}
\label{sec:moe_inference}

At inference time, \MethodLoRAMoE{}~and \MethodOLoRAMoE{} select an expert without access to the ground-truth task identity.
Given an embedding $\mathbf{e}$, each expert autoencoder $f_j$ produces a reconstruction loss
\begin{equation}
\label{eq:moe_recon_loss}
\ell_j(\mathbf{e}) = \lVert f_j(\mathbf{e}) - \mathbf{e} \rVert_2^2.
\end{equation}
Hard routing chooses the expert with minimal reconstruction loss
\begin{equation}
\label{eq:moe_select}
\hat{k}(\mathbf{e}) = \arg\min_{j\in\{0,\ldots,K-1\}} \ell_j(\mathbf{e}),
\end{equation}
and activates only one corresponding adapter for the task ($\hat{k}$).
Here $K$ denotes the number of experts trained so far (i.e., after stage $s$, $K=s$ in a $T$-task stream).
\paragraph{Unseen detection and fallback.}
If the input is classified as unseen, i.e., $\ell_j(\mathbf{e}) > \tau_j$ for all active experts $j$, we fall back to the direction-specific base adapter (as described in Section \ref{sec:transfer_motionllm}).
This prevents routing to an arbitrarily bad expert for out-of-distribution inputs and is used consistently in our evaluations.



\subsection{\MethodLoRAMoE{}-$K$ and \MethodOLoRAMoE{}-$K$ (Top$K$ mixture, $K>1$)}
\label{sec:olora_moe_m}

\MethodLoRAMoE{}-$K$ and \MethodOLoRAMoE{}-$K$ extend their hard-routing counterparts (Section~\ref{sec:olora_moe}) by allowing multiple experts to contribute at inference time.
The expert-training and router-training procedures remain unchanged.
After learning $s$ tasks, $s$ experts are available, and inference selects the best $\min(K,s)$ experts according to the router scores, with one autoencoder router per expert.
Below, Top$K$ denotes this selection capped by the number of available experts.
The routing is inspired by \citep{dmole2025}.

\paragraph{Router scores and Top$K$ selection.}
Given a routing embedding $\mathbf{e}\in\mathbb{R}^{H}$, each expert autoencoder $f_j$ yields a reconstruction loss $\ell_j(\mathbf{e})$, like in equation~\ref{eq:moe_recon_loss}.
We convert losses into routing logits by negating them,
\begin{equation}
\label{eq:olora_moe_m_logits}
 z_j(\mathbf{e}) = -\ell_j(\mathbf{e}),
\end{equation}
select the Top$K$ experts with largest logits (equivalently, smallest losses), and compute mixture weights by a softmax restricted to that Top$K$ set:
\begin{equation}
\label{eq:olora_moe_m_weights}
\omega_j(\mathbf{e}) = \frac{\exp(z_j(\mathbf{e}))}{\sum_{i\in\mathrm{Top}K(\mathbf{z}(\mathbf{e}))}\exp(z_i(\mathbf{e}))},
\qquad j\in\mathrm{Top}K(\mathbf{z}(\mathbf{e})).
\end{equation}
Setting $K=1$ recovers hard routing (equation~\ref{eq:moe_select}).


\paragraph{Logit-mixture soft routing.}
For the Top$K$ variants reported in the main results, soft routing is implemented by mixing output logits.
The routing weights are computed once per input and held fixed during autoregressive decoding.
Concretely, at decoding step $t$, each selected expert adapter $j\in\mathrm{Top}K(\mathbf{z}(\mathbf{e}))$ produces a next-token logit (score) vector $\mathbf{s}^{(j)}_t\in\mathbb{R}^{|\mathcal{V}|}$, where $\mathcal{V}$ is the cardinality of the token vocabulary.
We then form a weighted logit mixture
\begin{equation}
\label{eq:olora_moe_m_logitmix}
\mathbf{s}_t \,=\, \sum_{j\in\mathrm{Top}K(\mathbf{z}(\mathbf{e}))} \omega_j(\mathbf{e})\; \mathbf{s}^{(j)}_t,
\end{equation}
followed by greedy decoding of the next token-id ($y_t$), formally 
$$y_t = \underset{v \in \mathcal{V}}\arg\max {\mathbf{s}_t[v]},$$ where $\mathbf{s}_t[v]$ is the logit assigned to the token $v$. 
Importantly, this procedure maintains a separate key-value cache for each expert, i.e., each expert follows its own hidden-state trajectory and the mixture is performed only at the output-logit level.

An alternative is to combine the selected experts in LoRA update space, rather than mixing their output logits.
Appendix~\ref{sec:appendix_soft_routing_merge} derives unweighted and router-weighted merged adapters and compares their inference behavior with hard routing and logit-mixture decoding.
The construction applies to both expert families, while the comparative evaluation reported there is restricted to \MethodLoRAMoE{} without orthogonality regularization.

\subsection{Multi-task (upper bound)}
\label{sec:multitask}
Multi-task trains a single adapter on the union of all task data.
Since all tasks are observed throughout training, it does not suffer from catastrophic forgetting by construction and serves as an upper-bound reference.

%

%% file: sections/results.tex
\section{Experimental setup and results}
\label{sec:experiments}
The above-described approaches were evaluated on the five-task benchmark created from the HumanML3D dataset \citep{guo2022humanml3d}. The benchmark is described in Table \ref{tab:task_definitions}. It's construction is explained in Appendix \ref{sec:benchmark_construction}. 
The evaluation metrics are summarized in Section \ref{sec:metrics}, and the results are reported in Section \ref{sec:results}.

Note, our benchmark uses a balanced subset of approximately 5,700 samples from the HumanML3D dataset obtained through semantic clustering and overlap analysis to define five incremental tasks.
The original MotionLLM model in \cite{wu_motion-agent_2024} was trained on all 14,616 HumanML3D motion samples, so that the results reported there differ from our ({\bf Multi-task}) results.

The evaluation followed a two-tiered approach, starting with token accuracy to establish a baseline for CL performance, followed by qualitative generation metrics to assess the final output quality for T2M and M2T.

\begin{table}[t]
\caption{Continual benchmark task definitions (semantic clustering) with balanced splits (total: 5,700 samples).}
\label{tab:task_definitions}
\begin{center}
\begin{tabular}{cllccc}
\multicolumn{1}{c}{\bf Task} & \multicolumn{1}{c}{\bf Category} & \multicolumn{1}{c}{\bf Keywords} & \multicolumn{1}{c}{\bf Train} & \multicolumn{1}{c}{\bf Val} & \multicolumn{1}{c}{\bf Test}\\
\hline\\
1 & Running & runs, jogs, forward & 800 & 170 & 170\\
2 & Arms/Hands & left, right, hand, raises & 800 & 170 & 170\\
3 & Walking & walks, steps, turns & 800 & 170 & 170\\
4 & Jumping/Kicking & jumps, kicks, jumping & 800 & 170 & 170\\
5 & Sit/Stand & sits, stands, knees & 800 & 170 & 170\\
\hline
\end{tabular}
\end{center}
\end{table}

\subsection{Evaluation metrics}
\label{sec:metrics}
In the following, the different evaluation metrics are described. These comprise standard CL summary statistics on token accuracy as well as established metrics for motion and language generation quality.

\paragraph{Token accuracy and CL summary statistics.}
Let $s\in\{1,\ldots,T\}$ denote the training stage after learning tasks $1$ to $s$, and $\mathcal{T}_t, t\in\{1,\ldots,T\}$ denote the evaluation task.
For each CL method and direction, we built a performance matrix $R\in\mathbb{R}^{T\times T}$ with entries
\begin{equation}
\label{eq:perf_matrix}
R[s,t] := \mathrm{TokAcc}(\mathcal{T}_t;\;\text{model at stage }s),
\end{equation}
where $\mathrm{TokAcc}$ is the teacher-forced next-token accuracy on the output sequence (motion tokens for T2M, text tokens for M2T).
Based on the performance matrices, we computed accuracy (ACC), backward transfer (BWT), and forward transfer (FWT) for each direction and each approach as:
\begin{align}
\mathrm{ACC} &= \frac{1}{T}\sum_{t=1}^{T} R_{T,t}, \\
\mathrm{BWT} &= \frac{1}{T-1}\sum_{t=1}^{T-1}\bigl(R_{T,t}-R_{t,t}\bigr),\\
\mathrm{FWT} &= \frac{1}{T-1}\sum_{t=2}^{T} R_{t-1,t}.
\end{align}

\paragraph{Generation quality.}
For T2M, we report final-stage motion generation metrics: Fr\'{e}chet Inception Distance (FID), R-Precision at $K\in\{1,3\}$ (R@1, R@3), Diversity and motion-text matching distance (MM-Dist).
For M2T, we report final-stage caption quality: BLEU-1, BLEU-4, ROUGE-L, CIDEr, and BERTScore.

For an interpretation of these metrics, see Appendix \ref{sec:metric_interpretation}.

\subsection{Experimental Results}
\label{sec:results}
Note, the computational setup and hyperparamters used for the experiments are summarized in Appendix \ref{sec:comp_setup_hyperparams}.
\subsubsection{Token accuracy}
Table~\ref{tab:cl_metrics_bidir} reports ACC, BWT and FWT for both directions (T2M, M2T).
Orthogonal adapter hard routing (\MethodOLoRAMoE{}) yields near-zero forgetting (T2M BWT close to $0$ percentage points), while Transfer and O-LoRA-PM show substantial negative BWT. The newly combined joined MoE variants show that \textsc{O-LoRA-MoE-K2} yields the highest T2M ACC/FWT and the strongest M2T BWT, while \MethodOLoRAMoE{} remains best on M2T ACC/FWT and achieves the strongest T2M BWT. Notably, both \MethodOLoRAMoE{} and \textsc{O-LoRA-MoE-K2} outperform Multi-task in T2M token accuracy. Token accuracies for \MethodLoRAMoE{}-$K$ and \MethodOLoRAMoE{}-$K$, for $K=2,3,4,5$ can be found in \ref{sec:ablationTopK}. Tables regarding the percentage of fall-back to the base model, for method \MethodOLoRAMoE{}, can be found in \ref{sec:ablationFallback}.

\begin{table}[htbp]
\caption{Bidirectional continual learning metrics (token accuracy). ACC/FWT are in percent; BWT is in percentage points (pp). Multi-task is final-only and thus has no BWT/FWT.}
\label{tab:cl_metrics_bidir}
\begin{center}
\begin{tabular}{lcccccc}
\multicolumn{1}{c}{\bf Method} & \multicolumn{1}{c}{\bf T2M ACC} & \multicolumn{1}{c}{\bf M2T ACC} & \multicolumn{1}{c}{\bf T2M BWT} & \multicolumn{1}{c}{\bf M2T BWT} & \multicolumn{1}{c}{\bf T2M FWT} & \multicolumn{1}{c}{\bf M2T FWT}\\
\hline\\
Multi-task & 23.94 & 59.17 & -- & -- & -- & --\\
\hline
Transfer & 20.33 & 50.32 & $-$5.18 & $-$8.15 & 17.90 & 42.88\\
O-LoRA-PM & 19.94 & 50.30 & $-$6.00 & $-$9.04 & 18.49 & 42.30\\
 &  &  &  &  &  & \\
\textsc{LoRA-MoE} & 24.00 & 48.82 & $-$4.42 & $-$0.15 & 17.93 & 40.36\\
\textsc{LoRA-MoE-K2} & 30.18 & 51.78 & $-$0.57 & +0.99 & 21.79 & 43.17\\
\textsc{O-LoRA-MoE-K2} & \textbf{31.59} & 51.59 & $-$0.55 & \textbf{+2.19} & \textbf{22.24} & 42.47\\
\MethodOLoRAMoE{} & 27.68 & \textbf{54.01} & \textbf{+0.07} & $-$0.32 & 20.71 & \textbf{50.50}\\
\hline
\end{tabular}
\end{center}
\end{table}

\subsubsection{Motion generation quality}
Table~\ref{tab:t2m_quality} compares final-stage T2M generation quality.
Among the CL approaches, \MethodOLoRAMoE{} scores best on FID, Diversity, and MM-Dist, while \MethodLoRAMoE{} and \textsc{O-LoRA-MoE} achieve the strongest retrieval scores (R@1/R@3). \MethodOLoRAMoE{} even outperforms Multi-task in Diversity; however, it performs worse than Multi-task in FID, R@1, R@3 and MM-Dist.

\begin{table}[htbp]
\caption{T2M motion quality metrics (final stage, averaged across 5 tasks). $\downarrow$ = lower is better; $\uparrow$ = higher is better.}
\label{tab:t2m_quality}
\begin{center}
\begin{tabular}{lccccc}
\multicolumn{1}{c}{\bf Method} & \multicolumn{1}{c}{\bf FID $\downarrow$} & \multicolumn{1}{c}{\bf R@1 $\uparrow$} & \multicolumn{1}{c}{\bf R@3 $\uparrow$} & \multicolumn{1}{c}{\bf Diversity $\uparrow$} & \multicolumn{1}{c}{\bf MM-Dist $\downarrow$}\\
\hline\\
Multi-task  & 1.39 & 10.3\% & 22.9\% & 6.57 & 2.97\\
\hline
Transfer & 45.25 & 3.2\% & 7.1\% & 6.06 & 7.66\\
O-LoRA-PM & 34.82 & 3.0\% & 7.7\% & 6.95 & 7.20\\
 &  &  &  &  & \\
\textsc{LoRA-MoE} & 16.79 & \textbf{6.25\%} & 13.53\% & 6.46 & 6.09\\
\textsc{LoRA-MoE-K2} & 61.41 & 3.75\% & 10.65\% & 3.45 & 8.58\\
\textsc{O-LoRA-MoE-K2} & 60.63 & 4.35\% & 11.33\% & 3.18 & 8.33\\
\MethodOLoRAMoE{} & \textbf{10.39} & 6.0\% & \textbf{14.3\%} & \textbf{7.19} & \textbf{4.70}\\
\hline
\end{tabular}
\end{center}
\end{table}


\subsubsection{Motion caption quality}
Table~\ref{tab:m2t_nlg} reports final-stage M2T natural language generation (NLG) quality.
Among the CL methods, \textsc{LoRA-MoE} achieves the strongest overall caption quality, outperforming Multi-task on all reported NLG metrics. Note, the corresponding \textsc{LoRA-MoE-K2} and \textsc{O-LoRA-MoE-K2} variants show substantially degraded captioning quality in this configuration.

\begin{table}[htbp]
\caption{M2T NLG quality metrics (final stage, averaged across 5 tasks). Higher is better.}
\label{tab:m2t_nlg}
\begin{center}
\begin{tabular}{lccccc}
\multicolumn{1}{c}{\bf Method} & \multicolumn{1}{c}{\bf BLEU-1} & \multicolumn{1}{c}{\bf BLEU-4} & \multicolumn{1}{c}{\bf ROUGE-L} & \multicolumn{1}{c}{\bf CIDEr} & \multicolumn{1}{c}{\bf BERTScore}\\
\hline\\
Multi-task  & 41.47 & 9.32 & 34.86 & 14.63 & 86.00\\
\hline
Transfer & 23.39 & 2.23 & 25.17 & 3.60 & 84.20\\
O-LoRA-PM & 26.78 & 2.39 & 27.88 & 5.54 & 84.89\\
 &  &  &  &  & \\
\textsc{LoRA-MoE} & \textbf{45.93} & \textbf{11.26} & \textbf{37.15} & \textbf{30.77} & \textbf{88.96}\\
\textsc{LoRA-MoE-K2} & 1.57 & 0.04 & 3.98 & 0.30 & 76.08\\
\textsc{O-LoRA-MoE-K2} & 1.57 & 0.06 & 4.02 & 0.53 & 76.10\\
\MethodOLoRAMoE{} & 32.57 & 8.22 & 36.43 & 21.52 & 88.81\\
\hline
\end{tabular}
\end{center}
\end{table}

\subsubsection{Router accuracy}
Table~\ref{tab:routing_accuracy} reports \MethodOLoRAMoE{} per-sample routing accuracy at the final stage.
Despite imperfect per-sample routing, hard selection of a single expert is sufficient for strong end-to-end performance, as shown in the previous results.
The results for \MethodLoRAMoE{} are similar.
\begin{table}[htbp]
\caption{\MethodOLoRAMoE{} per-sample routing accuracy at the final stage (all five tasks learned).}
\label{tab:routing_accuracy}
\begin{center}
\begin{tabular}{lcc}
\multicolumn{1}{c}{\bf Task} & \multicolumn{1}{c}{\bf T2M (\%)} & \multicolumn{1}{c}{\bf M2T (\%)}\\
\hline
Running & 83.1 & 83.1\\
Arms/Hands & 83.1 & 83.8\\
Walking & 66.2 & 79.9\\
Jumping/Kicking & 77.3 & 77.3\\
Sit/Stand & 67.5 & 67.5\\
\hline
\textbf{Average} & \textbf{75.5} & \textbf{78.3}\\
\hline
\end{tabular}
\end{center}
\end{table}

%% file: sections/discussion.tex




\section{Discussion}
\label{sec:discussion}
Beyond performance metrics, our design choices—integrating LoRA, orthogonality, and MoE-style routing—are functionally tailored for embodied skill acquisition, where preserving expert isolation prevents interference between motion representations while the router enables the dynamic selection of competencies required by varying environmental instructions.
Our results demonstrate that  low-rank adapters with routed experts offer a practical way for transforming a static motion-language model such as Motion-Agent \citep{wu_motion-agent_2024} into an incrementally learning motion-language agent.
Across both directions (M2T and T2M),  subspace isolation substantially reduces catastrophic forgetting compared to sequential fine-tuning and adapter merging.
The bidirectional continual metrics (Table~\ref{tab:cl_metrics_bidir}) show a  stability hierarchy. \MethodOLoRAMoE{} achieves near-zero forgetting (BWT close to 0 percentage points) in both directions, whereas Transfer and O-LoRA-PM incur substantial negative BWT. 
This is consistent with the stability-plasticity dilemma. Methods that modify a shared parameterization across tasks (Transfer, Progressive Merge) enable adaptation but can overwrite prior task competence. The orthogonality regularization can have a positive effect on token accuracy and BWT.
The results also highlight a consistent directional asymmetry. Token accuracy is considerably higher in M2T than in T2M for every method.
This suggests that, under this benchmark and model configuration, mapping motion tokens to language is an easier prediction problem than generating motion tokens from text. Moreover, orthogonal regularization seems to be less beneficial for M2T, where \MethodLoRAMoE{} consistently outperforms \MethodOLoRAMoE{} on the NLG metrics. 

For deployment, task identity is typically unavailable.
\MethodOLoRAMoE{} addresses this by routing inputs to a single orthogonal expert (hard Top{1} selection).
While per-sample routing accuracy is imperfect (Table~\ref{tab:routing_accuracy}), the end-to-end performance remains competitive. 
Notably, routing accuracy varies by task (e.g., lower on Walking and Sit/Stand), indicating that some motion categories are less separable under the chosen routing representations.

We also observed that Top$K$ soft routing with mixture decoding (logit averaging for $K>1$) is substantially less stable than hard routing ($K=1$) for motion captioning.
For the Top$K$ variants in the main results, soft routing is realized via logit-mixture decoding with separate key-value caches per expert (equation~\ref{eq:olora_moe_m_logitmix}).
This means experts follow different hidden-state trajectories while their next-token logits are combined at each step.
One possible explanation for the poor caption quality is that expert disagreement leads to token choices that individual experts assign low probability, with errors compounding during autoregressive decoding and degrading sequence-level metrics (BLEU, ROUGE, CIDEr).
Moreover, router scores derived from reconstruction losses are not calibrated as blending coefficients.
The divergence between teacher-forced token accuracy and poor generation performance is consistent with this hypothesis, but does not establish the cause of the degradation.
The merged-adapter comparison in Appendix~\ref{sec:appendix_soft_routing_merge} further shows that generation quality depends on the inference realization in the evaluated non-orthogonal configuration.
However, differences in M2T decoding strategy prevent attributing the observed performance gaps solely to the location of expert mixing.


Finally, the results indicate that token-level continual metrics do not fully characterize downstream behavior.
E.g. Multi-task achieves the best M2T token accuracy (Table~\ref{tab:cl_metrics_bidir}), while \MethodLoRAMoE{} and \MethodOLoRAMoE{} outperform Multi-task on some of the text generation quality metrics (Table~\ref{tab:m2t_nlg}, ROUGE-L, CIDEr, and BERTScore).
This divergence motivates evaluation protocols that report both token-level CL metrics and generation-quality metrics when studying continual motion-language agents.


%% file: sections/conclusion.tex
\section{Conclusion}
\label{sec:conclusion}
This work studied parameter-efficient continual learning for bidirectional motion--language agents that must both caption motion (M2T) and generate motion from language (T2M) under sequential task exposure.
Building on Motion-Agent \citep{wu_motion-agent_2024}, we evaluated a family of adapter-based continual methods on a five-task HumanML3D benchmark derived via semantic clustering of text descriptions.

Our results support three main conclusions.
First, orthogonal adapter isolation substantially mitigates forgetting: \MethodOLoRAMoE{} achieves near-zero backward transfer (BWT close to 0 percentage points) in both directions while maintaining competitive final accuracy.
Second, task-label-free inference via routing is feasible without oracle task identity: Although \MethodOLoRAMoE{} routes correctly only in $76,9\%$ of samples on average (Table~\ref{tab:routing_accuracy}), hard expert selection is sufficient to obtain competitive end-to-end performance in this benchmark. 
Third, routing strategy matters: Compared to mixture routing (using logit-mixture) of $K>1$ experts, hard Top1 routing over experts yields better downstream generation quality, indicating that preserving expert consistency might be preferable.
Finally, we observe that token-level continual metrics do not fully predict downstream generation quality.
This motivates bidirectional evaluation protocols that report both CL summary statistics and task-relevant quality metrics.
\paragraph{Limitations and future work.}
Our study is limited to five tasks derived from semantic clustering of motion descriptions (which can result in kinematic overlapping) with a fixed task order determined by greedy farthest-point selection on cluster centroids (to maximize inter-task diversity) and to a specific router representation (mean-pooled input embeddings). It therefore has a narrow scope and provides an initial analysis of catastrophic forgetting, parameter-efficient adaptation, and task-agnostic routing in motion-language continual learning, a setting that has not previously been explored systematically. Moreover, the benchmark should be viewed as an initial, interpretable continual-learning setup rather than a definitive decomposition of motion domains.
Future work should, however, examine longer task horizons and alternative task definitions (e.g., kinematic or interaction-centric tasks), should improve task-label-free routing by incorporating richer routing signals (e.g., learned task descriptors) and evaluate alternative soft expert blending techniques. 
Moreover, a comprehensive task-order sensitivity study should be conducted to evaluate the influence of task-order.
More broadly, continual objectives that better align with generation quality - rather than token accuracy alone - may be necessary.

\newpage

%% file: sections/appendix.tex
\appendix
\section{Appendix}
\label{sec:appendix}
\subsection{Additional details of \MethodOLoRAMoE}
\label{sec:overview_fig}

The \MethodOLoRAMoE{} architecture illustrated in Figure~\ref{fig:overview} consists of task-incremental training and task-agnostic inference, detailed below.
\begin{enumerate}
\item \textbf{Task-Incremental Training (Sequential Style Learning):} The model processes a continuous stream of motion tasks ($T_1 \dots T_n$) using a dual-phase training protocol for each new task $T_k$.\\
    \textbf{Phase 1}: Expert Training ($\phi_k$): A task-specific LoRA expert ($A_k, B_k$) is trained with an orthogonality regularizer against prior experts $j < k$. This encourages reduced overlap between task-specific low-rank subspaces to mitigate inter-task interference. \\
    \textbf{Phase 2}: Router Training ($f_k$): After the expert is fixed, a lightweight MLP autoencoder is trained specifically for that task. It learns to reconstruct mean-pooled LLM input embeddings ($e$) for task $T_k$, using text-prompt embeddings for T2M and motion-token embeddings for M2T. These embeddings have 2304 dimensions for Gemma-2-2B.
\item \textbf{Task-Agnostic Inference (Adaptive Style Selection):} During deployment, the model processes inputs without explicit task labels.\\
    \textbf{Reconstruction Check:} The LLM input embeddings are mean-pooled to obtain $e$, which is then evaluated by all trained routers ($f_1 \dots f_n$). Each router produces a reconstruction loss $\ell_j(e) = \|f_j(e) - e\|^2_2$.\\
    Expert Selection (Winner-Takes-All): The system identifies the expert $\hat{k}$ corresponding to the router with the minimal reconstruction loss.\\
    \textbf{Hard Routing Decision:} If $\ell_j(e) > \tau_j$ for all routers, where $\tau_j = \mu_j + 2\sigma_j$ is the threshold estimated from router $j$'s training losses, the system treats the input as unseen and falls back to the direction-specific base adapter. Otherwise, it activates the expert $\hat{k}$ with the minimal reconstruction loss.
\end{enumerate}
\textbf{The Dual-Role Backbone:} The system integrates a Frozen Foundation Backbone (Gemma-2-2B LLM and VQ-VAE tokenizer/decoder) with two modular components. The backbone serves a dual purpose:
\begin{enumerate}
    \item[a)] 
    It acts as the stable structural base that is extended by the Expert Bank (modular O-LoRA adapters) to provide task-specific knowledge for both motion-to-text (M2T) and text-to-motion (T2M) directions.\\
    \item[b)] 
    Its input embedding layer provides the token embeddings that are mean-pooled to train and operate the Router Module (parallel autoencoders).
\end{enumerate}
\subsection{Benchmark construction and task stream}
\label{sec:benchmark_construction}

Tasks were constructed by semantic clustering of HumanML3D's text descriptions as follows:
\begin{itemize}
    \item Text preprocessing and embedding extraction: Each HumanML3D motion sample was associated with 1-3 natural-language descriptions. For clustering, we concatenated all descriptions per sample into a single text string and computed a single embedding using GTE-large (1024 dimensions)~\citep{li2023generaltextembeddingsmultistage}.
    \item Hierarchical clustering and task balancing: We applied hierarchical agglomerative clustering with Ward's minimum-variance linkage to the embedding vectors. The number of clusters was selected by scanning $k \in [5,100]$ and evaluating silhouette score and Davies-Bouldin index. Both criteria selected $k{=}5$. To obtain balanced continual tasks, we subsampled each cluster to 1,140 samples and (deterministically) splitted each task into 800/170/170 train/validation/test.
    \item Pretraining split (10\%): In addition to the five continual tasks, we reserved 10\% of the total (872 samples) as a pretraining subset used to construct the pretrained model checkpoint (cf. Section \ref{sec:transfer_motionllm}). The remaining samples were used for continual learning such that the sequential methods do not train on any held-out pretraining samples.
    \item Task ordering and evaluation split convention: Tasks were presented in a fixed order determined by greedy farthest-point selection on cluster centroids (to maximize inter-task diversity). For training we followed the \emph{trainval} convention and evaluated exclusively on the held-out test split.
\end{itemize}

\subsection{Further explanation of task loss $\mathcal{L}_{\text{task}}$}
\label{sec:taskloss_explained}
The components of the task loss, as defined in  \eqref{eq:transfer_task_loss}, as
\begin{equation}
\mathcal{L}_{\text{task}}^{(t)}
= - \frac{1}{|\Omega|}\sum_{i\in\Omega} \log p_{\theta,\phi}(z_i\mid z_{<i})
\end{equation}
are defined as follows:
\paragraph{Token log-likelihood term.}
The probability $p_{\theta,\phi}(z_i\mid z_{<i})$ is the model's next-token distribution
at position $i$ under teacher forcing.
Let $\mathbf{s}_i\in\mathbb{R}^{|\mathcal{V}|}$ denote the decoder logits over the extended
vocabulary $\mathcal{V}$ produced after consuming the prefix $z_{<i}$.
Then
\begin{equation}
p_{\theta,\phi}(z_i\mid z_{<i}) = \mathrm{softmax}(\mathbf{s}_i)[z_i],
\end{equation}
and $\log p_{\theta,\phi}(z_i\mid z_{<i})$ is the log-likelihood assigned to the ground-truth
token $z_i$ (equivalently, the negative cross-entropy contribution for that position).

\paragraph{Direction-specific definition of $\Omega$ (T2M vs. M2T).} All instruction and input tokens are
masked and therefore excluded from $\Omega$.
Only the response region contributes to the loss.

\textbf{T2M:}
Given a caption $\mathbf{c}=(c_1,\ldots,c_{N})$ and a discrete motion code sequence
$\mathbf{m}=(m_1,\ldots,m_{L})$ obtained from the frozen VQ-VAE tokenizer, the
training sequence has the form
\begin{equation}
\mathbf{z}_{\text{T2M}} = [\texttt{<bos>},\; \text{prompt}(\mathbf{c}),\; \texttt{<Motion>},\; \mathbf{m},\; \texttt{</Motion>},\; \texttt{<eos>}],
\end{equation}
where \texttt{<bos>} and \texttt{<eos>} denote beginning and end of sentence. All tokens up to and including the opening delimiter \texttt{<Motion>} are prompt tokens
and are masked.
Thus, the supervised index set is
\begin{equation}
\Omega_{\text{T2M}} = \{ i : z_i \in [\mathbf{m},\; \texttt{</Motion>},\; \texttt{<eos>}] \},
\end{equation}
and the task loss becomes
\begin{equation}
\mathcal{L}_{\text{T2M}}^{(t)}
= -\frac{1}{|\Omega_{\text{T2M}}|} \sum_{i\in\Omega_{\text{T2M}}} \log p_{\theta,\phi}(z_i\mid z_{<i}).
\end{equation}

\textbf{M2T:}
Given a motion code sequence $\mathbf{m}$ and its caption $\mathbf{c}$, the training
sequence is
\begin{equation}
\mathbf{z}_{\text{M2T}} = [\texttt{<bos>},\; \text{prompt}(\mathbf{m}),\; \texttt{<Motion>},\; \mathbf{m},\; \texttt{</Motion>},\; \mathbf{c},\; \texttt{<eos>}],
\end{equation}
where the prompt (including the entire motion span inside \texttt{<Motion>}\dots\texttt{</Motion>})
is masked.
Therefore, the supervised index set is
\begin{equation}
\Omega_{\text{M2T}} = \{ i : z_i \in [\mathbf{c},\; \texttt{<eos>}] \},
\end{equation}
and the task loss is
\begin{equation}
\mathcal{L}_{\text{M2T}}^{(t)}
= -\frac{1}{|\Omega_{\text{M2T}}|} \sum_{i\in\Omega_{\text{M2T}}} \log p_{\theta,\phi}(z_i\mid z_{<i}).
\end{equation}

\subsection{Interpretation of evaluation metrics}
\label{sec:metric_interpretation}

\paragraph{Token accuracy and CL statistics}
Teacher-forced token accuracy provides a stable and comparable scalar for both directions, enabling us to populate the full matrix $R[s,t]$ across stages $s$ and tasks $t$.
ACC summarizes final average competence after learning all (previous) tasks, BWT measures retention vs. forgetting of earlier tasks (with values close to 0 percentage points indicating minimal forgetting), and FWT measures forward transfer to future tasks before they are explicitly learned.

\paragraph{Motion generation quality.}
FID evaluates how closely the distribution of generated motions matches that of real motions in a learned feature space, providing a global proxy for realism and coverage.
R@1, R@3 (R-Precision) assesses text-motion alignment by testing whether the ground-truth text is retrieved among the top-$K$ candidates given a generated motion (higher is better).
MM-Dist further quantifies embedding-space alignment between generated motion and its conditioning text (lower is better), complementing the discrete top-$K$ retrieval view.
Diversity measures the variability of generated motions and is used to detect mode collapse, ensuring that continual adaptation does not preserve quality by degenerating to overly similar outputs.

\paragraph{Text generation quality.}
BLEU-1 and BLEU-4 capture n-gram precision at coarse and strict granularities, respectively, reflecting surface-level correctness of generated descriptions.
ROUGE-L measures longest-common-subsequence overlap and is sensitive to sentence-level content coverage.
CIDEr is designed for captioning with multiple references, rewarding consensus with the reference set and better reflecting that many captions can be valid for the same motion.
BERTScore measures semantic similarity in a contextual embedding space and is included to account for paraphrases that may be penalized by purely n-gram-based metrics.

\subsection{Computational setup and hyperparameters}
\label{sec:comp_setup_hyperparams}
All experiments ran on a single machine with $4\times$ NVIDIA L40S GPUs (48\,GB VRAM each), 192\,GB RAM, and an AMD EPYC CPU.
Multi-GPU training used HuggingFace Accelerate with Distributed Data Parallel (DDP) and bfloat16 mixed precision.

The hyperparameters shared across all methods are summarized in Table \ref{tab:shared_hyperparams}.
\begin{table}[htbp]
\caption{Shared training hyperparameters across all methods.}
\label{tab:shared_hyperparams}
\begin{center}
\begin{tabular}{ll}
\multicolumn{1}{c}{\bf Parameter} & \multicolumn{1}{c}{\bf Value}\\
\hline\\
Learning rate & $10^{-4}$\\
Epochs per task & 20\\
Micro-batch size & 8 per GPU\\
Gradient accumulation & 4 steps\\
Effective batch size & $8\times 4\times 4\text{ GPUs} = 128$\\
Optimiser & AdamW ($\beta_1{=}0.9$, $\beta_2{=}0.999$)\\
Mixed precision & bfloat16\\
\hline
\end{tabular}
\end{center}
\end{table}
The LoRA adapters targeted the attention projection layers with dropout 0.05, rank/alpha 64/64 for T2M and 32/32 for M2T, and  an orthogonality penalty with $\lambda{=}0.5$.
For routing-based methods, the autoencoder router operated on mean-pooled LLM input embeddings (2304 dimensions for Gemma-2-2B) with hidden dimension 512 and was trained for 300 epochs per task with learning rate $10^{-3}$.

\subsection{Ablation study expert selection $K>1$}
\label{sec:ablationTopK}
We ablate the number of selected experts $K$ in the joined MoE variants.
Here, $K{=}1$ corresponds to hard (winner-takes-all) expert selection, while $K{>}1$ blends the top$K$ experts.
Table~\ref{tab:cl_metrics_bidir_k_sweep} isolates the effect of $K$ for \textsc{LoRA-MoE} and \textsc{O-LoRA-MoE}.


\begin{table}[htbp]
\caption{Joined MoE ablation: effect of selecting the top-$K$ experts on bidirectional continual learning metrics (token accuracy). ACC/FWT are in percent; BWT is in percentage points (pp).}

\label{tab:cl_metrics_bidir_k_sweep}
\begin{center}
\begin{tabular}{lcccccc}
\multicolumn{1}{c}{\bf Method} & \multicolumn{1}{c}{\bf T2M ACC} & \multicolumn{1}{c}{\bf M2T ACC} & \multicolumn{1}{c}{\bf T2M BWT} & \multicolumn{1}{c}{\bf M2T BWT} & \multicolumn{1}{c}{\bf T2M FWT} & \multicolumn{1}{c}{\bf M2T FWT}\\
\hline\\
\textsc{LoRA-MoE-K2} & 30.18 & \textbf{51.78} & $-$0.57 & +0.99 & 21.79 & 43.17\\
\textsc{LoRA-MoE-K3} & 31.31 & 49.70 & \textbf{+0.45} & $-$0.65 & 19.64 & \textbf{44.22}\\
\textsc{LoRA-MoE-K4} & 30.02 & 47.56 & +0.10 & $-$3.11 & 21.31 & 43.88\\
\textsc{LoRA-MoE-K5} & 28.78 & 48.67 & $-$1.08 & $-$2.20 & 21.47 & 43.12\\
\hline
\textsc{O-LoRA-MoE-K2} & 31.59 & 51.59 & $-$0.55 & \textbf{+2.19} & 22.24 & 42.47\\
\textsc{O-LoRA-MoE-K3} & \textbf{31.79} & 49.45 & $-$0.25 & $-$0.67 & 23.99 & 43.97\\
\textsc{O-LoRA-MoE-K4} & 31.45 & 47.97 & $-$1.49 & $-$1.62 & \textbf{25.00} & 43.31\\
\textsc{O-LoRA-MoE-K5} & 31.04 & 47.71 & $-$1.45 & $-$2.92 & 24.89 & 44.06\\
\hline
\end{tabular}
\end{center}
\end{table}

Table~\ref{tab:cl_metrics_bidir_k_sweep} shows that moving from $K{=}1$ (hard selection) to a small amount of blending ($K{=}2$ or $3$) substantially improves T2M accuracy for both LoRA-MoE and O-LoRA-MoE (e.g., O-LoRA-MoE increases from 23.16\% to 31.59\%/31.79\%).
However, larger $K$ offers diminishing returns and can harm retention: several settings exhibit more negative BWT at $K{=}4$ or $5$ (especially in M2T), consistent with soft blending reducing expert isolation.
Overall, the best trade-off is achieved at small $K$: O-LoRA-MoE-$K{=}2$ yields the strongest M2T retention (BWT $+2.19$ pp), while O-LoRA-MoE-$K{=}3$ yields the strongest T2M ACC (31.79\%).

\input{sections/soft_routing_with_merged_adapter}

\subsection{Fallback rate on seen tasks (O-LoRA-MoE).}
\label{sec:ablationFallback}

In the task-agnostic routing setup of O-LoRA-MoE, the router selects a task expert (LoRA adapter) for each input.
If the router cannot confidently map a sample to any of the seen task experts, it routes the sample to a generic fallback expert (here: the direction adapter for M2T or for T2M).
We report the \emph{fallback\_rate\_on\_seen}, defined for a seen task as the fraction of evaluation samples that are routed to the fallback expert.
Lower values indicate more reliable expert selection.
For stages where a task has not been learned yet, the metric is undefined and is shown as $-$.

\begin{table}[h]
\centering
\small
\begin{tabular}{lccccc}
\hline
Stage & Jumping & Arms/Hands & Walking & Gestures & Sit/Stand \\
\hline
Task 0 & 5.9 & --  & --  & --  & --  \\
Task 1 & 6.9 & 9.9 & --  & --  & --  \\
Task 2 & 6.9 & 9.9 & 2.0 & --  & --  \\
Task 3 & 0.0 & 9.9 & 1.0 & 6.9 & --  \\
Task 4 & 0.0 & 7.9 & 1.0 & 6.9 & 4.0 \\
\hline
\end{tabular}
\caption{Fallback rate on seen tasks (\%, lower is better) for O-LoRA-MoE on the M2T stream.}
\label{tab:olora-moe-fallback-on-seen-m2t}
\end{table}

\begin{table}[h]
\centering
\small
\begin{tabular}{lccccc}
\hline
Stage & Jumping & Arms/Hands & Walking & Gestures & Sit/Stand \\
\hline
Task 0 & 5.9 & --  & --  & --  & --  \\
Task 1 & 5.9 & 5.9 & --  & --  & --  \\
Task 2 & 5.0 & 7.9 & 9.9 & --  & --  \\
Task 3 & 0.0 & 6.9 & 2.0 & 4.0 & --  \\
Task 4 & 3.0 & 5.9 & 8.9 & 5.0 & 4.0 \\
\hline
\end{tabular}
\caption{Fallback rate on seen tasks (\%, lower is better) for O-LoRA-MoE  on the T2M stream.}
\label{tab:olora-moe-fallback-on-seen-t2m}
\end{table}

%% file: sections/soft_routing_with_merged_adapter.tex
\subsection{\texorpdfstring{Top$K$}{Top-K} soft routing with merged adapters (weight-space mixture)}
\label{sec:appendix_soft_routing_merge}

Section~\ref{sec:olora_moe_m} describes Top$K$ soft routing through output-logit mixing (equation~\ref{eq:olora_moe_m_logitmix}).
Here, we study a distinct inference alternative that combines the selected LoRA updates into a single adapter.
The construction does not require orthogonality between experts and therefore applies to both \MethodLoRAMoE{} and \MethodOLoRAMoE{}.
The empirical comparison below is restricted to \MethodLoRAMoE{} ($\lambda_{\mathrm{orth}}=0$); it does not evaluate the effect of orthogonality regularization on merged-adapter inference.
Because LoRA updates enter each adapted linear projection additively, we can define a weight-space mixture of the selected expert updates.
For any adapted linear projection with frozen weight $W_0$ and a column-vector input activation $x\in\mathbb{R}^{d_{\text{in}}}$, the Top$K$ mixture applies
\begin{equation}
\label{eq:olora_moe_m_layer}
W(x)
= W_0 x
+ \sum_{j\in\mathrm{Top}K(\mathbf{z}(\mathbf{e}))} \omega_j(\mathbf{e})\; \tfrac{\alpha}{r}\, B_j A_j x,
\end{equation}
where $(A_j,B_j)$ are the LoRA factors of expert $j$, and $\omega_j(\mathbf{e})$ are the router-provided mixture weights.

\paragraph{Unweighted merge.}
Because LoRA deltas add linearly, for any adapted projection the expert-specific deltas are
\(\Delta W_j = \tfrac{\alpha}{r} B_j A_j\).
For a selected expert set \(\mathcal{S}=\{j_1,\ldots,j_K\}\), the sum \(\Delta W_{\mathrm{sum}}=\sum_{j\in\mathcal{S}}\Delta W_j\) can be implemented exactly by constructing a new LoRA adapter with rank $r_{\mathrm{merge}}=Kr$ and concatenating factors:
\begin{equation}
\label{eq:appendix_merged_concat}
A_{\mathrm{merge}}=\mathrm{cat}(A_{j_1},\ldots,A_{j_K}),\qquad
B_{\mathrm{merge}}=\mathrm{cat}(B_{j_1},\ldots,B_{j_K}).
\end{equation}
Here, $A$ factors are concatenated along their rows and $B$ factors along their columns.
We set $\alpha_{\mathrm{merge}}=K\alpha$, preserving $\alpha_{\mathrm{merge}}/r_{\mathrm{merge}}=\alpha/r$ for experts with the same rank and scaling.
This yields $B_{\mathrm{merge}}A_{\mathrm{merge}}=\sum_{m=1}^{K}B_{j_m}A_{j_m}$, so the merged adapter represents the exact sum of deltas, not their average.

\paragraph{Router-weighted merge.}
To match the weighted mixture in equation~\ref{eq:olora_moe_m_layer}, we scale the $B$ factors before concatenation:
\begin{equation}
\label{eq:appendix_merged_weighted}
A_{\mathrm{merge}}=\mathrm{cat}(A_{j_1},\ldots,A_{j_K}),\qquad
B_{\mathrm{merge}}=\mathrm{cat}(\omega_{j_1}(\mathbf{e})B_{j_1},\ldots,\omega_{j_K}(\mathbf{e})B_{j_K}).
\end{equation}
This yields $\Delta W_{\mathrm{merge}}=\sum_{j\in\mathcal{S}(\mathbf{e})}\omega_j(\mathbf{e})\,\Delta W_j$.
The router weights are softmax-normalized over the selected experts, so $\sum_{j\in\mathcal{S}(\mathbf{e})}\omega_j(\mathbf{e})=1$.
The implementation reuses a temporary merged adapter, overwriting its factors for each input because the weights are input-dependent; unweighted adapters can instead be cached by the selected expert set.
Both constructions preserve the original LoRA scaling as above.

\paragraph{Relation to logit-mixture decoding.}
The merged-adapter formulation applies the mixture inside each adapted linear layer and follows a single model trajectory per decoding hypothesis, without maintaining separate expert key--value caches.
However, experts here are low-rank deltas on a shared backbone (rather than independent subnetworks), and routing weights are computed once per input and held fixed during decoding.
Exact equivalence at the adapted linear layers does not imply equivalence to mixing the output logits of separate nonlinear models.

\paragraph{Experimental comparison.}
We compare hard routing ($K=1$), logit-mixture decoding ($K=2$), and unweighted and router-weighted merging ($K=2$) using the \MethodLoRAMoE{} expert and router checkpoints without orthogonality regularization ($\lambda_{\mathrm{orth}}=0$).
Within each direction, all four variants reuse the same trained expert and router checkpoints at each stage; there is no merge-specific retraining.
The runs use five sequential tasks, 20 training epochs per task, and the random 80/20 split with split seed 42.
Final-stage generation metrics are macro-averaged over five tasks, with 101 evaluation samples per task (505 in total).
The source experts have rank 32 for M2T and 64 for T2M, giving merged ranks of 64 and 128, respectively, when two experts are selected.
Before two experts are available, $K$ is capped by the number of active experts; samples rejected by the router use the direction-specific fallback adapter.
These are single-run comparisons, not averages over independent training seeds.

\begin{table}[htbp]
\centering
\small
\caption{M2T caption quality for the \MethodLoRAMoE{} inference variants ($\lambda_{\mathrm{orth}}=0$), macro-averaged over five tasks at the final stage. Higher is better; NLG scores use the evaluator's $100\times$ reporting scale. Hard routing and merges use two-beam decoding; logit mixing uses greedy decoding.}
\label{tab:merged_routing_m2t}
\begin{tabular}{lrrrrr}
\hline
Inference & BLEU-1 & BLEU-4 & ROUGE-L & CIDEr & BERTScore\\
\hline
Hard routing ($K=1$) & \textbf{45.93} & \textbf{11.26} & \textbf{37.15} & \textbf{30.77} & \textbf{88.96}\\
Logit mixture ($K=2$) & 1.57 & 0.04 & 3.98 & 0.30 & 76.08\\
Unweighted merge ($K=2$) & 26.69 & 3.92 & 27.40 & 10.06 & 86.56\\
Weighted merge ($K=2$) & 44.63 & 9.40 & 34.85 & 17.46 & 88.77\\
\hline
\end{tabular}
\end{table}

\paragraph{Captioning results.}
Table~\ref{tab:merged_routing_m2t} shows that weighted merging improves every reported caption-quality metric over unweighted merging.
BLEU-4 increases from 3.92 to 9.40, ROUGE-L from 27.40 to 34.85, and CIDEr from 10.06 to 17.46.
Nevertheless, hard routing remains stronger on all five metrics.
In particular, the relatively close BLEU-1 and BERTScore values should not obscure the remaining CIDEr gap: weighted merging scores 17.46 versus 30.77 for hard routing, approximately 43.3\% lower.
Thus, weighting improves the merged-adapter realization for M2T in this run, but does not establish an advantage over selecting a single expert.

\begin{table}[htbp]
\centering
\small
\caption{T2M generation quality for the same inference comparison ($\lambda_{\mathrm{orth}}=0$), macro-averaged over five tasks at the final stage. R@1/R@3 are the stored retrieval fractions multiplied by 100. Bold indicates the best value in each column.}
\label{tab:merged_routing_t2m}
\begin{tabular}{lrrrrr}
\hline
Inference & FID $\downarrow$ & R@1 (\%) $\uparrow$ & R@3 (\%) $\uparrow$ & Diversity $\uparrow$ & MM-Dist $\downarrow$\\
\hline
Hard routing ($K=1$) & \textbf{16.79} & \textbf{24.95} & \textbf{54.06} & \textbf{6.46} & \textbf{6.09}\\
Logit mixture ($K=2$) & 61.41 & 15.05 & 42.57 & 3.45 & 8.58\\
Unweighted merge ($K=2$) & 21.92 & 19.80 & 48.71 & 6.37 & 6.81\\
Weighted merge ($K=2$) & 34.98 & 17.23 & 48.51 & 4.76 & 7.48\\
\hline
\end{tabular}
\end{table}

\paragraph{Motion generation and metric disagreement.}
The T2M results exhibit a different ordering (Table~\ref{tab:merged_routing_t2m}).
Unweighted merging achieves lower FID than weighted merging (21.92 versus 34.98), higher diversity (6.37 versus 4.76), and lower motion--text matching distance (6.81 versus 7.48); it also has higher R@1 and R@3.
Hard routing remains strongest on all five generation metrics shown.
Conversely, weighted merging achieves higher teacher-forced token accuracy than unweighted merging (29.63\% versus 24.09\%; Table~\ref{tab:merged_routing_cl}).
Thus, improved token-level prediction does not necessarily translate into improved generated motion, and the preferred merging rule depends on the generation direction.

\begin{table}[htbp]
\centering
\small
\caption{Token-accuracy CL metrics for the \MethodLoRAMoE{} inference variants ($\lambda_{\mathrm{orth}}=0$). ACC/FWT are percentages; BWT is in percentage points. Metrics follow the definitions in Section~\ref{sec:metrics} and use all five stages.}
\label{tab:merged_routing_cl}
\begin{tabular}{lrrrrrr}
\hline
& \multicolumn{3}{c}{M2T} & \multicolumn{3}{c}{T2M}\\
Inference & ACC & BWT & FWT & ACC & BWT & FWT\\
\hline
Hard routing ($K=1$) & 48.82 & $-0.15$ & 40.36 & 24.00 & $-4.42$ & 17.93\\
Logit mixture ($K=2$) & 51.78 & $+0.99$ & 43.17 & 30.18 & $-0.57$ & 21.79\\
Unweighted merge ($K=2$) & 41.53 & $-3.32$ & 37.17 & 24.09 & $-2.78$ & 17.70\\
Weighted merge ($K=2$) & 46.54 & $-0.69$ & 39.49 & 29.63 & $-1.13$ & 22.22\\
\hline
\end{tabular}
\end{table}

\paragraph{Interpretation and limitations.}
Weighted and unweighted merging differ in both expert weighting and update magnitude: the unweighted variant sums the selected updates, whereas the weighted variant uses normalized router weights.
Consequently, this comparison does not isolate router-informed weighting from update normalization; a uniform-average merge would provide an additional control.
Both merges improve the reported generation metrics relative to the evaluated logit-mixture baseline, but this comparison concerns implemented inference pipelines rather than a fully controlled isolation of mixture location.
In particular, M2T logit mixing uses greedy decoding, whereas hard routing and merged adapters use two-beam decoding.
The observed gains therefore cannot be attributed solely to weight-space merging.
The single-model decoding formulation avoids separate expert caches, but a higher merged rank and per-input adapter construction also incur costs; these runs provide no measured latency or memory comparison.
No repeated-seed uncertainty estimates were used, so the comparisons are descriptive rather than claims of statistical significance. \\
Within the evaluated non-orthogonal configuration, router-weighted merging yields better M2T caption quality than unweighted merging, whereas unweighted merging yields better T2M generation metrics.
Hard routing achieves the best generation scores among the four compared inference variants.